\documentclass[letterpaper, 10 pt, conference]{ieeeconf}  
\usepackage[utf8]{inputenc}
\usepackage{amsmath, amsfonts}
\usepackage{graphicx}
\usepackage{xcolor}
\usepackage{subcaption}
\usepackage{wrapfig}
\usepackage{adjustbox}
\usepackage{xspace}
\usepackage{epstopdf}
\usepackage{xcolor}
\usepackage{float}
\usepackage{soul}
\usepackage{hyperref}
\usepackage{siunitx}
\usepackage{booktabs}
\usepackage{algorithm}
\usepackage{algcompatible}
\usepackage[capitalize]{cleveref}
\usepackage{booktabs}
\usepackage{multirow}
\usepackage[para]{threeparttable} 
\usepackage{array}
\usepackage{colortbl}
\usepackage{siunitx}
\usepackage{pifont}

\usepackage{eqexpl}
\eqexplSetDelim{=}

\usepackage{algpseudocode}
\usepackage{array}
\usepackage{tabularx}

\usepackage{wrapfig}
\usepackage{caption}
\usepackage{subcaption}
\usepackage[noadjust]{cite}
\usepackage{comment}

\IEEEoverridecommandlockouts 

\newcommand{\systemname}{\textit{}\xspace}

\title{\systemname Egocentric Tactile and Proximity Sensors as Observation Priors for Humanoid Collision Avoidance}

\author{Carson Kohlbrenner, Niraj Pudasaini, William Xie, Naren Sivagnanadasan,\\Nikolaus Correll, and Alessandro Roncone
\thanks{All authors are with the University of Colorado Boulder, 1111 Engineering Drive, Boulder, CO USA. This work is partially supported by NSF FW-
HTF-R grant \#2222952.
{\tt\small name.surname@colorado.edu}.}}
\begin{document}
\maketitle

\begin{abstract}

Collision-free motion is often aided by tactile and proximity sensors distributed on the body of the robot due to their resistance to occlusion as opposed to external cameras. However, how to shape the sensor's properties, such as sensing coverage; type; and range, to enable avoidant behavior remains unclear. In this work, we present a reinforcement learning framework for whole-body collision avoidance on a humanoid H1-2 robot and use it to characterize how sensor properties shape learned avoidance behavior. Using dodgeball as a benchmark task, we ablate the properties of sensors distributed across the upper body of the robot and find that raw proximity measurements can substitute for explicit object localization provided the sensing range is sufficient and that sparse non-directional proximity signals outpace dense directional alternatives in sample efficiency.
\end{abstract}


\section{Introduction}

Humanoid robots operating in dynamic environments near humans, such as walking in crowded corridors or participating in sports, require preemptive avoidance for anticipated collisions for safe deployment \cite{kong2026learning, xue2026collision}. Whole-body artificial skins that sense contact, as well as actionable information near the robot for anticipating contact, are commonly used to aid in collision prevention as opposed to external cameras due to their resistance to occlusion, low observation complexity, and egocentric placement \cite{navarro2021proximity}. Classical control paradigms use constraints for how the robot should react to signals emanating from such artificial skin \cite{armleder2025real, murooka2024whole}, but learned policies using reinforcement learning (RL) use the sensor signals themselves to learn how to react \cite{miller2025enhancing}.

Sensors used for collision anticipation include time-of-flight (ToF) \cite{kim2024armor, escobedo2021contact, borelli2024generating}, capacitive \cite{kohlbrenner2026gentactprox}, and acoustic sensors \cite{fan2021aurasense} that each have distinct measurement principles that limit their sensing coverage and signal quality. Although various iterations of whole-body sensors for collision avoidance have been proposed, the desirable qualities of a proximity signal for collision avoidance remains unclear. For example, capacitive sensing skins can detect objects radially less than \qty{20}{cm} away with signal degradation over distance \cite{kohlbrenner2026gentactprox}, whereas ToF sensors can detect objects meters away along a directional measurement axis in a narrow field of view (FoV) \cite{kim2024armor}. How to shape the signals of such egocentric sensors to act as effective observation priors that couple the robot's body to its environment in a way that makes reactive behaviors more learnable remains largely unexplored.

In this work, we simulate an H1-2 humanoid robot with distributed egocentric tactile and proximity sensors on its upper body with a variety of sensing coverage geometries and signal qualities to identify what type of observation representation is desirable for learned collision compliant control. We isolate the observation priors imposed by the sensors as a representation bias for learning by providing no rewards on the sensor states. Our contributions are as follows: 1) A reinforcement learning framework that utilizes egocentric sensors to guide collision avoidant whole-body control on a humanoid robot, and 2) a comparison of how sensor signals impact learned collision avoidance behavior.

\begin{figure}
    \centering
    \includegraphics[width=0.7\linewidth]{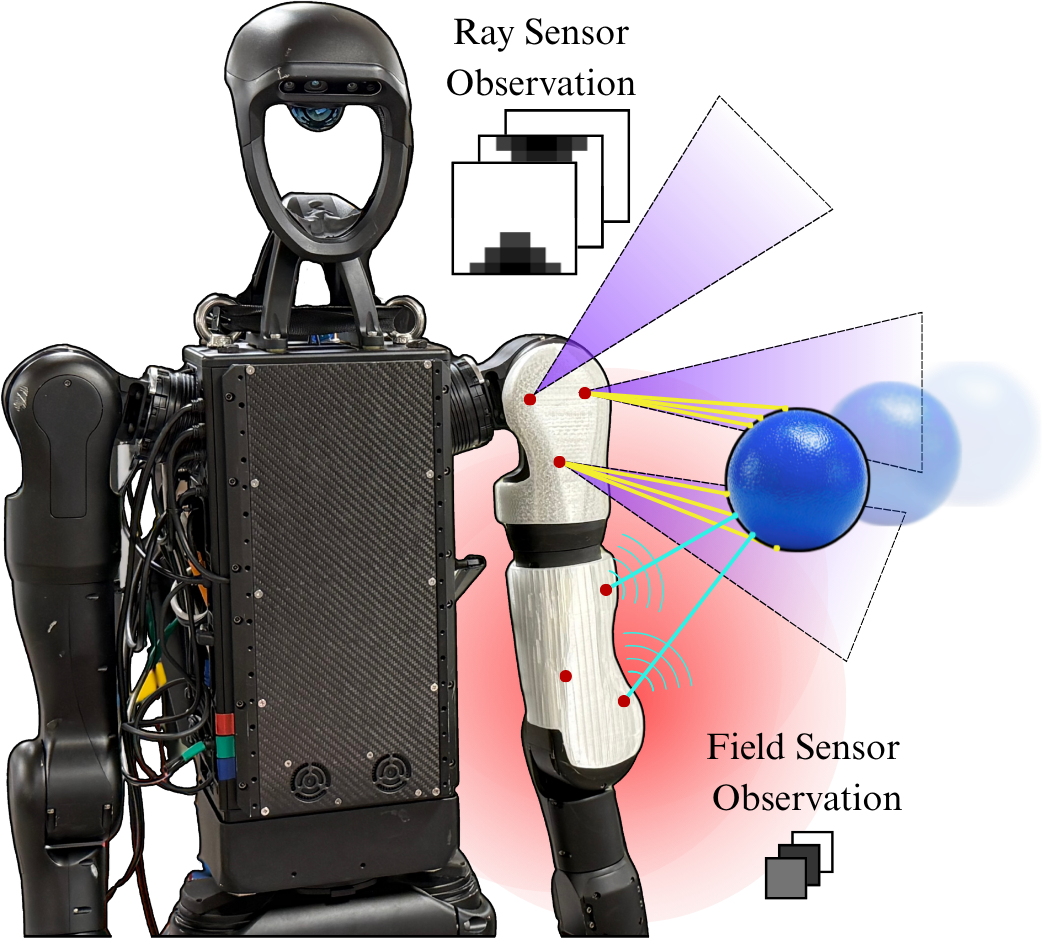}
    \caption{Distributed sensors reveal information about nearby objects, with yellow lines indicating ray sensor detections and blue lines indicating field sensor detections.}
    \label{fig:cover}
    \vspace{-0.4cm}
\end{figure}
\section{Methods}
To characterize the effects of egocentric distributed sensors, we simulate an H1-2 humanoid robot with 64 sensors distributed on the upper-half of the robot as seen in \cref{fig:sensor_types} playing an idealized form of dodgeball---a sport with the objective to dodge incoming thrown balls.

\subsection{Whole-Body Collision Avoidance}
The policy $\pi_\theta(a_t|o_t)$ outputs joint-level commands to maximize the expected cumulative reward:
\begin{equation}
\max_\theta \ \mathbb{E}\left[\sum_{t=0}^{T} \gamma^t r_t(o_t, a_t)\right],
\end{equation}
where $\gamma$ is the discount factor and $r_t$ encodes both stability and safety. Training is conducted using the Proximal Policy Optimization (PPO) \cite{schulman2017proximalpolicyoptimizationalgorithms} algorithm, chosen for its robustness and scalability to high-dimensional continuous control. The policy outputs target joint positions $q^*_t \in \mathbb{R}^{21}$. These are converted into motor torques $\tau_t$ via a low-level PD controller:
\begin{equation}
\tau_t = K_p (q^*_t - q_t) + K_d (\dot{q}^*_t - \dot{q}_t),
\end{equation}
where $K_p$ and $K_d$ are the proportional and derivative gains, and $\dot{q}^*_t$ is approximated by zero or finite differences. This action formulation favors compliant, stable motion compared to direct torque control. Next, we define the collision avoidance task's rewards, observations, and termination conditions.

\textbf{Rewards:} We formulate the reward $r_t$ as a weighted sum of \textit{task objective} (i.e. survival time) and \textit{regularization terms} (i.e. energy efficiency, posture, and drift). While the task reward drives the robot to discover agile dodging maneuvers to survive, the regularization terms are critical for suppressing high-frequency jitter and ensuring the learned behaviors are fluid, energy-efficient, and natural. In this work, we do not provide rewards for maintaining clearance from the object as to not favor a specific sensor modality.

\textbf{Observations:} The observation space is comprised of base angle velocity, relative joint positions, relative joint velocities, the gravity vector projected in the base frame to estimate torso orientation and stability, and the distributed sensor signals. Additionally, we adopt an asymmetric actor-critic structure \cite{pinto2017asymmetricactorcriticimagebased} to enhance training stability while ensuring deployability, where the critic is provided with a privileged linear base velocity observation. The actor and critic are each two-layer feedforward networks with 32 units per layer. This fixed, low-capacity architecture was deliberately chosen so that differences in task performance can be attributed to the sensor morphology rather than model expressivity, treating the body's distributed sensing geometry as an inductive bias that fundamentally shapes what behaviors are learnable \cite{bengio2013representation}.

\subsection{Sensors}

A variety of configurations of distributed tactile and proximity sensors are simulated to evaluate their impacts on collision avoidant behavior learning. For a network of $S$ sensors, the signal $\mathcal{O}^i$ of sensor $i$ at time $t$ is at a minimum the function $f_i$ of the detected object's shape $V_{\text{obj}}$, position $p_{\text{obj}}^i$, and orientation $q_{\text{obj}}^i$ in the sensor's reference frame:
\begin{equation}
    \mathcal{O}^i(t)=f_i(V_{\text{obj}}, p_{\text{obj}}^i(t), q_{\text{obj}}^i(t)|\theta_i)
\end{equation}
where $\theta_i$ is the set of sensor $i$'s intrinsic parameters that describe its signal. To isolate the best intrinsic parameters $\theta$, we retrain the policy with a variety of sensing coverage geometries, signal types, and detection ranges.

\begin{figure}
    \centering
    \includegraphics[width=\linewidth]{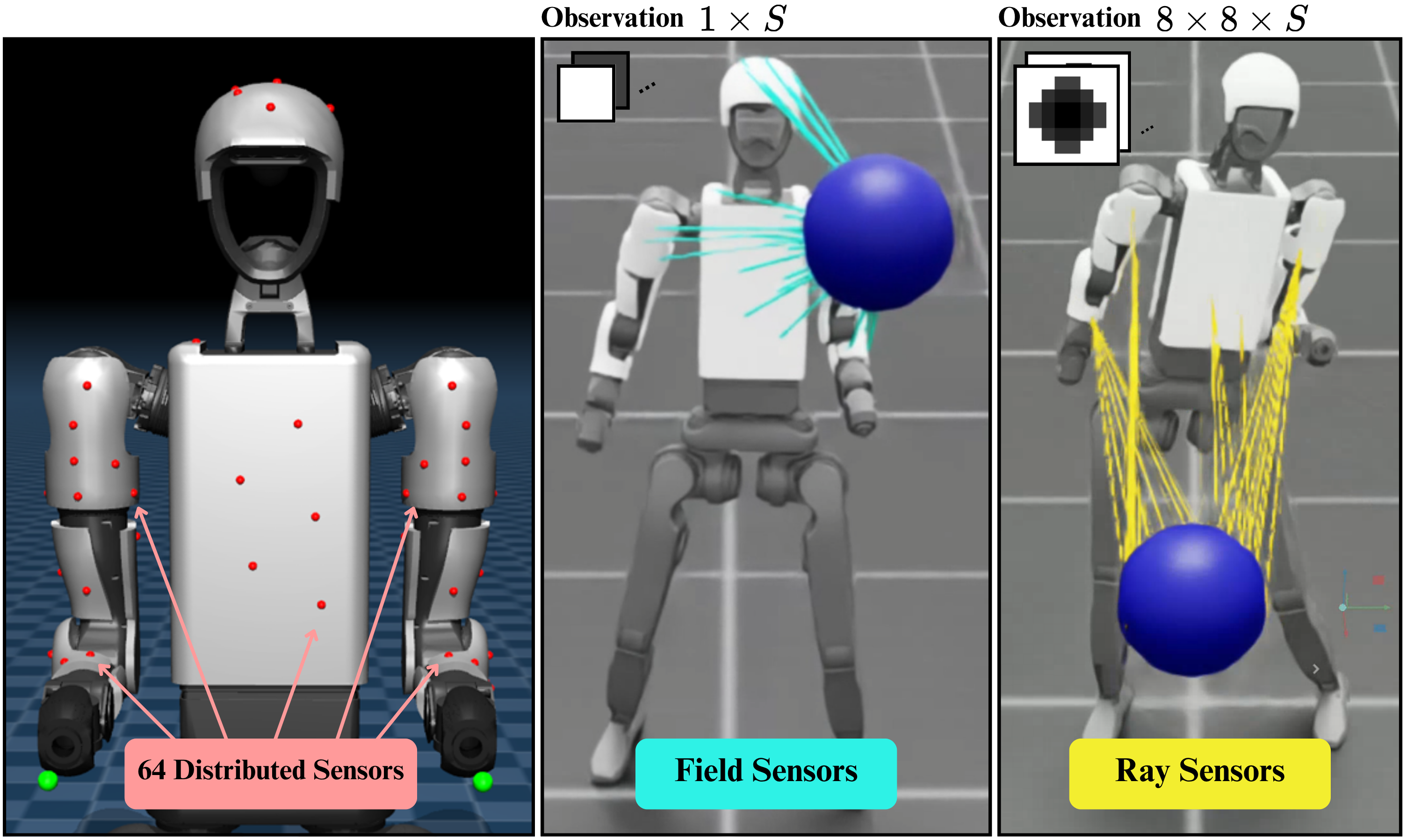}
    \caption{The colored lines connecting the ball to the robot indicate a distance measurement from a given sensor, where ray sensors are restricted to a set direction and field sensors return any distance within range.}
    \label{fig:sensor_types}
    \vspace{-0.2 cm}
\end{figure}

\textbf{Proximity Sensing:} The simulated coverage geometries and signal properties are \textit{sensor-agnostic} abstractions inspired by real proximity sensors. The tested coverage geometries consist of spherical fields inspired by capacitive and acoustic proximity sensors \cite{rupavatharam2023ambisense, kohlbrenner2026gentactprox} and rays inspired by time-of-flight (ToF) proximity sensors \cite{kim2024armor}. The field sensors return a non-directional signal if an object is within its receptive field, whereas ray sensor return a signal if an object intersects with its directional beam. Each ray sensor is given an 8x8 grid of beams with $63^{\circ}$ square diagonal field of view (FoV) (shown in \cref{fig:sensor_types}) to emulate contemporary ToF sensing works \cite{kim2024armor}. The observation space of ray sensors is thus $z_{ray}^{8\times8\times S}$ and the field sensor observation space is $z_{field}^{S}$.

The tested signal functions $f$ include perfect localization of the object, relative distance measurements (proximity), and binary detection. The perfect localization signal returns the position of the ball in cartesian coordinates relative to the robot's center frame if \textit{any} of the sensors can detect the ball. The perfect localization signal serves as an upper bound for training, and having multiple sensors observe the ball with this signal function does not reveal more information to the agent. The relative distance signal function returns the distance of the closest point on the ball in range of the sensor normalized by the sensor's maximum sensing range. The binary detection signal function returns a boolean that is true if the ball is within range of the sensor.

The sensor positions on the robot were instantiated randomly using the GenTact design pipeline \cite{kohlbrenner2024gentact}. The sensors are idealized, meaning their positions on the robot are known and there is no noise, latency, or measurement failures to test the impacts of the \textit{desirable} signal on collision avoidance learning. Additionally, all objects besides the incoming balls are segmented out of a sensor's signals (including the robot itself \cite{borelli2024generating}) to isolate the spatial relationship between the robot and incoming objects in the signal (\cref{fig:sensor_types}).

\textbf{Tactile Sensing:} Although the dodgeball task is designed to reward reactive collision free motion, we propose that tactile sensors can be used for reward shaping through data segmentation to identify when trajectories are collision free. Binary contact detection tactile sensors are distributed on each link of the robot to provide full-body coverage. When any of the tactile sensors are activated during training, the episodes are terminated and the agent can no longer collect rewards for staying upright and alive.

\section{Results}

\begin{figure*}
    \centering
    \includegraphics[width=\linewidth]{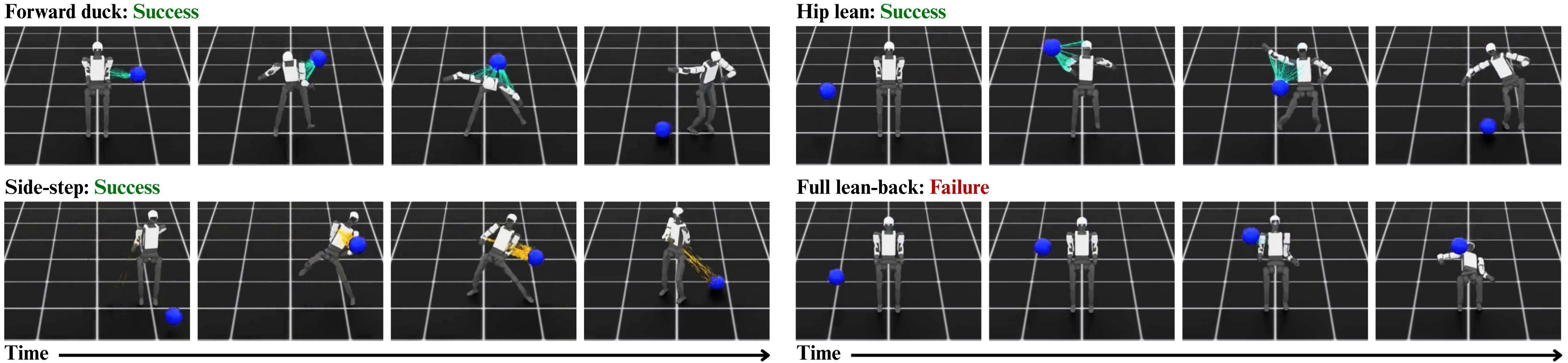}
    \caption{Common policies learned for the dodgeball collision avoidance task. Policies that kept the robot standing yielded higher rewards than those that caused the robot to fall, even if the ball was successfully dodged.}
    \label{fig:sequence}
\end{figure*}

\begin{figure*}
    \centering
    \includegraphics[width=\linewidth]{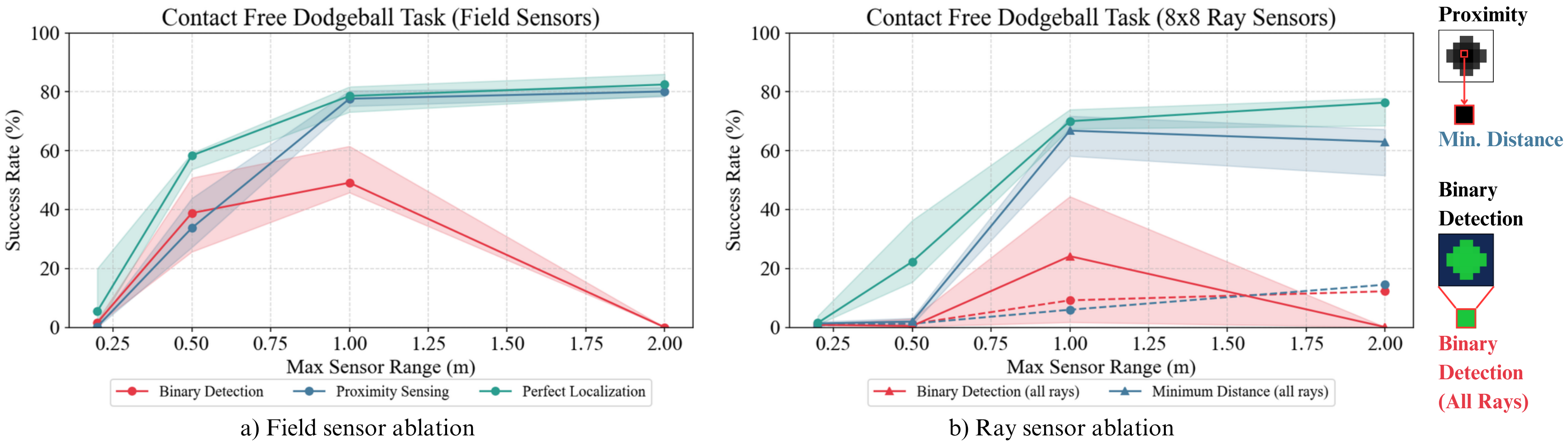}
    \caption{Ablation of sensor coverage geometry and signal type for training the H1-2 how to avoid collisions in dodgeball. The reported line is the average of the IQR for ten uniquely seeded and retrained policies for each point.}
    \label{fig:both}
    \vspace{-0.2cm}
\end{figure*}

\subsection{Experimental Setup}
 At the start of each episode, a ball with a 15~cm radius is thrown at the robot, followed by additional balls thrown every 1–2 seconds. Balls are spawned 2–3 m above the ground and 4–6 m from the robot, with an initial velocity of 4–8 m/s. All ranges are sampled uniformly. This environment rewards sustained upright contact free posture, and success is determined if the robot can avoid contact for three seconds.

\subsection{Sensor Ablation}

A batch of 4096 agents was trained in parallel for 3000 epochs per permutation of sensor coverage geometry, signal type, and maximum range (0.2, 0.5, 1, and 2 m). Ten seeded models were trained for each permutation to identify variance in the learned policies, where each run was trained for a maximum of 910 simulated hours. \cref{fig:sequence} shows some of the commonly learned policies that led to success or failure. The top and bottom 25\% of runs were excluded to get the interquartile range (IQR). The IQR and its mean shown in \cref{fig:both} reveal three primary findings: 1) \textit{relative proximity} measurements served as strong alternatives to localization at high sensing ranges for both ray and field sensors, 2) binary detection signals served as strong alternatives to localization signals at low sensing ranges, and 3) sparse non-directional proximity signals improved sample efficiency compared to dense directional signals.



In line with intuition, perfect localization yields the strongest performance across all sensing ranges for both coverage geometries. Perfect localization with the ray sensor performed noticeably worse than the field sensor, especially at small sensing ranges, which may be attributed to the ray sensors having more blind spots than the field sensors. 

Surprisingly, the field and ray sensors diverged significantly in performance for both relative proximity and binary detection measurements. \cref{fig:both}-a shows that relative proximity measurements provide as strong an observational bias as perfect localization at high sensing ranges, but relative proximity performs very poorly compared to perfect localization in \cref{fig:both}-b. We attribute this finding to the high \textit{signal density} of ray sensors, each of which returns 64 distance measurements as a depth image. To test this finding, we retrained the agent with ray proximity sensors to instead return the minimum distance of all its beams, shaping its observation dimension to be the same as the field sensor. This agent substantially outperformed the full depth image variant and achieved peak performance nearly identical to that of the perfect localization signal type at a 1~m range.

Finally, binary detection outperformed relative distance measurements for field sensors at low sensing ranges (0.5~m), shown in \cref{fig:both}-a, and the reduced ray sensor that returned a positive detection if any beam detected the ball had a slight performance gain at 1~m over the full ray detection sensor. One explanation may be that simply knowing the ball is nearby is enough information for reactive control, and having a denser signal with distance information hinders the agent's search through the solution space.

\subsection{Discussion}



These findings have implications for how distributed sensors should be deployed practically. Perfect localization represents privileged information that can be estimated through sensor fusion techniques such as Kalman filtering \cite{khodarahmi2023review}. Skipping the state estimation step entirely and passing in relative proximity measurements to the agent may be a cheaper alternative depending on the technical setup that our findings suggest may be just as practical given a large enough sensing range. This may be ideal if the localization step adds substantial latency or if the location of nearby objects cannot reliably be estimated to a single Cartesian coordinate, such as a robot navigating a crowded room \cite{zou2023object}. 

One objective of shaping the signals of onboard sensors is improve to reduce the amount of data needed to learn the task. We noticed that the only configurations that failed to converge on average within the 3000 epoch limit were the ray-based proximity and binary sensor variants, whereas higher performing runs such as the field-based proximity sensors at ranges greater than 1~m converged to stable policies in under 1000 epochs which trains in $\approx \qty{20}{min}$ on a consumer grade RTX 4090 GPU or $\approx300$ simulated hours of experience. This result suggests that properly shaping the signals of the distributed sensors can lead to a substantial reduction in data and may be necessary for practical deployment on a real humanoid robot.




\section{Conclusion}

We introduced a reinforcement learning framework for whole-body collision avoidance on a humanoid robot and used it to characterize how egocentric sensor signal properties shape learned avoidance behavior. Our ablation across coverage geometry, signal type, and sensing range reveals that raw proximity measurements can substitute for explicit object localization, provided the sensing range is sufficient. At short ranges where proximity information is sparse, binary detection signals prove viable, suggesting that coarse spatial awareness is sufficient for reactive whole-body motion. We further find that signal density is a critical and underappreciated design consideration: sparse non-directional proximity signals consistently outpaced dense directional alternatives in sample efficiency. Together, these findings demonstrating that simpler, lower-bandwidth sensing configurations in an ideal scenario can achieve collision avoidance performance comparable to sensors with far richer output, and that tactile contact detection provides a natural mechanism for reward shaping without additional supervision.

\subsection{Limitations and Future Steps}
The findings in the work introduce further questions regarding ideal signal shaping and real-world deployment that should be addressed in future steps. Regarding ideal signal shaping, the ideal signal space should be further explored to allow the agent to observe time-sequences of observations, which may allow the agent to better infer the ball's direction at a given moment. Regarding real-world deployment, the noise profile of a sensor is heavily dependent on the sensing modality used. For example, capacitive sensors and acoustic sensors were both classified as field sensors in this work, but a capacitive sensor reading may be influenced by nearby electrostatic fields whereas acoustic sensors may be influenced by sound in the environment. Future steps should include profiling how the specific modalities noise can impact training. Additionally, existing whole-body artificial skin sensors should be used to validate the findings of this work on a real-world deployment \cite{kohlbrenner2026gentactprox, kim2024armor}.

\bibliographystyle{IEEEtran}
\bibliography{references}
\end{document}